\documentclass[sigconf]{acmart}

\AtBeginDocument{%
  \providecommand\BibTeX{{%
    \normalfont B\kern-0.5em{\scshape i\kern-0.25em b}\kern-0.8em\TeX}}}

\acmConference[RLEM '20]{RLEM '20: First Int'l Workshop on Reinforcement Learning for Energy Management in Buildings \& Cities}{November 03--05, 2020}{Yokohama, JP}

\usepackage{siunitx}
\begin{document}

\title[Centralised SAC Deep RL Approach to District Demand Side Management]{A Centralised Soft Actor Critic Deep Reinforcement Learning Approach to District Demand Side Management through CityLearn}

\author{Anjukan Kathirgamanathan}
\email{anjukan.kathirgamanathan@ucdconnect.ie}
\orcid{0003-0125-5235}
\authornotemark[1]
\affiliation{%
  \institution{School of Mechanical and Materials Engineering, University College Dublin}
  \city{Dublin}
  \country{Ireland}
}

\author{Kacper Twardowski}
\email{kacper.twardowski@ucdconnect.ie}

\author{Eleni Mangina}
\email{eleni.mangina@ucd.ie}
\affiliation{%
  \institution{School of Computer Science, University College Dublin}
  \streetaddress{P.O. Box 1212}
  \city{Dublin}
  \postcode{43017-6221}
}

\author{Donal P. Finn}
\affiliation{%
  \institution{School of Mechanical and Materials Engineering, University College Dublin}
  \city{Dublin}
  \country{Ireland}
}

\renewcommand{\shortauthors}{Kathirgamanathan, et al.}

\begin{abstract}
  Reinforcement learning is a promising model-free and adaptive controller for demand side management, as part of the future smart grid, at the district level. This paper presents the results of the algorithm that was submitted for the CityLearn Challenge, which was hosted in early 2020 with the aim of designing and tuning a reinforcement learning agent to flatten and smooth the aggregated curve of electrical demand of a district of diverse buildings. The proposed solution secured second place in the challenge using a centralised `Soft Actor Critic' deep reinforcement learning agent that was able to handle continuous action spaces. The controller was able to achieve an averaged score of 0.967 on the challenge dataset comprising of different buildings and climates. This highlights the potential application of deep reinforcement learning as a plug-and-play style controller, that is capable of handling different climates and a heterogenous building stock, for district demand side management of buildings.
\end{abstract}

\ccsdesc[500]{Applied computing~Engineering}
\ccsdesc[500]{Computing methodologies~Multi-agent reinforcement learning}

\keywords{Deep Reinforcement Learning, Smart Grid, Demand Side Management}


\maketitle

\section{Introduction}

Buildings are complex systems influenced by changing weather, occupancy, schedules and in a demand response (DR) context - grid signals. Capturing these dynamics in a physics-based simulation model capable of facilitating DR is highly challenging. This is due to highly non-linear behaviour of the thermal-dynamics and the fact that no one building is identical to another in the highly heterogenous building stock \cite{Kathirgamanathan2020}. Reinforcement learning (RL) is a model-free algorithm that learns from historical and real-time data and has shown promise in recent research applied to building energy management problems \cite{Vazquez-Canteli2019, Wang2020}. Given the novelty associated with RL in this domain, the behaviour of multiple energy consuming agents (i.e., buildings), subject to demand-dependent grid signals, is an area that is not well understood \cite{Vazquez-Canteli2019}. The CityLearn project (https://github.com/intelligent-environments-lab/CityLearn) is an OpenAI Gym environment \cite{1606.01540}, which aims to facilitate the implementation of RL agents in a multi-agent DR context for a diverse group of buildings \cite{Vazquez-Canteli2019a}. The main objective of CityLearn is to facilitate and standardize the evaluation and comparison of different RL agents and algorithms. \href{https://sites.google.com/view/citylearnchallenge}{The CityLearn Challenge}  was organised virtually and ran from January to July 2020. It invited participants to design, develop and tune a RL agent to flatten and smooth the aggregated curve of electrical demand for a district comprising of 9 diverse buildings. The current paper presents the results from a centralised "Soft Actor Critic" deep RL based algorithm that was submitted to the challenge. 

\section{Related Work}

The review of RL for DR by \citet{Vazquez-Canteli2019} shows its promising potential as a model-free technique, mitigating the need to develop physics-based control-oriented models, and capable of dealing with the heterogenous nature of the building stock. The review found that most studies to date focus on single building systems with demand-independent electricity prices. Focusing on deep RL, which has gained significant interest and traction in recent years, e.g., using Deep Q Networks \cite{Mnih2015}, such approaches have often been limited to discrete and low-dimensional action spaces \cite{Lillicrap2016}. There is a research gap in the application of deep reinforcement learning to problems with continuous action spaces in the building energy management domain.

The Soft Actor-Critic (SAC) algorithm, an off-policy maximum entropy actor-critic algorithm, as first proposed by Haarnoja et al \cite{Haarnoja2018} in 2018, is one of the algorithms that is capable of operating over continuous action spaces. At their core, actor-critic methods are a type of policy gradient method, which have separate memory structures to explicitly represent the policy \cite{SuttonRichardS.;Barton2014}. The policy structure is known as the actor and the estimated value function is known as the critic. The actor selects the actions whereas the critic evaluates the actions made by the actor. The reader is referred to Sutton \cite{SuttonRichardS.;Barton2014} for a more detailed explanation of Actor-Critic methods. \citet{Haarnoja2018} suggest that the SAC algorithm provides for both sample-efficient learning and stability and hence extends readily to complex, high-dimensional tasks. They found the SAC algorithm showed substantial improvement in both performance and sample efficiency over both off-policy and on-policy prior methods. This current research investigates the suitability of the SAC algorithm for tackling the district DSM problem utilising CityLearn.

\section{Methods}

\subsection{CityLearn Challenge}

The CityLearn challenge used a multi-objective cost function of five equally weighted metrics applied to an entire district of nine buildings (as outlined in Table \ref{table:1}). These are described below:
\begin{enumerate}
    \item Peak electricity demand (for the evaluation period of 1 year)
    \item Average daily electricity peak demand (daily peak demand of the district averaged over the evaluation period)
    \item Ramping (a measure of how much the district electricity consumption changes from one timestep to the next)
    \item 1 - Load factor (the average monthly electricity demand divided by its maximum peak)
    \item Net electricity consumption of the district over the evaluation period
\end{enumerate}
The multi-objective cost function is normalised by a baseline cost obtained from the performance of a predefined manually tuned Rule-Based Controller (RBC). This implied that a cost function of less than 1 resulted in a better performance than the RBC. This RBC controller charges cooling (and domestic hot water (DHW) if available) during the night and discharges during the day based only on the hour of the day. The adaptive potential of RL to deal with different environments, rather than solely the one it was trained on, was tested through evaluating the controller on different datasets. Participants used the design dataset to implement their RL agent (including design, tuning and pre-training) and could test their agent on the evaluation dataset and receive feedback through the cost function based on the generalisation results of their agents. Each dataset contains year-long hourly information about the cooling and DHW demand of the building, electricity consumed by appliances, solar power generation, as well as weather data and other variables. The evaluation dataset featured different buildings from different cities than the design dataset, albeit within the same climate zones (see Table \ref{table:2}). The challenge dataset is different from both the design and evaluation datasets featuring different buildings and climates.

\begin{table}[!htbp]
\small
\caption{Buildings and Descriptions in CityLearn Challenge District (Design Dataset)}
\label{table:1}
\centering
\begin{tabular}{p{1cm} c p{1.5cm} p{0.8cm} p{0.8cm} p{0.8cm}} 
 \hline
 Building Number & Type & Type Details & Cooling Storage$^1$ & DHW Storage$^1$ & PV (kW) \\ [0.5ex] 
 \hline
 1 & Commercial & Medium Office & 3 & 3 & 120 \\
 2 & Commercial & Fast-food Restaurant & 3 & 3 & N/A \\
 3 & Commercial & Standalone Retail & 3 & N/A & N/A \\
 4 & Commercial & Strip Mall Retail & 3 & N/A & 40 \\ [1ex] 
 5 & Residential & Medium Multi-family & 3 & 3 & 25 \\ [1ex] 
 6 & Residential & Medium Multi-family & 3 & 3 & 20 \\ [1ex] 
 7 & Residential & Medium Multi-family & 3 & 3 & N/A \\ [1ex] 
 8 & Residential & Medium Multi-family & 3 & 3 & N/A \\ [1ex] 
 9 & Residential & Medium Multi-family & 3 & 3 & N/A \\ [1ex] 
 \hline
\end{tabular}
\\
\emph{$^1$The storage capacity is the non-dimensional scaling factor given above multiplied by the building maximum cooling or DHW demand.}
\end{table}

\begin{table}[!htbp]
  \small
  \caption{Climate Zones for the Different Datasets in the CityLearn Challenge}
  \label{table:2}
  \centering
  \begin{tabular}{p{1.6cm}p{1.5cm}p{1.7cm}p{1.5cm}}
    \hline
    CityLearn Climate Zone & ASHRAE Identifier & Description & City \\ \hline
    \multicolumn{4}{c}{\textbf{Design Dataset}} \\
    \textbf{1} & 2A & Hot-Humid & New Orleans \\
    \textbf{2} & 3A & Warm-Humid & Atlanta \\
    \textbf{3} & 4A & Mixed-Humid & Nashville \\
    \textbf{4} & 5A & Cold-Humid & Chicago \\
    \hline
    \multicolumn{4}{c}{\textbf{Evaluation Dataset}} \\
    \textbf{1} & 2A & Hot-Humid & Orlando \\
    \textbf{2} & 3A &  Warm-Humid & Dallas \\
    \textbf{3} & 4A & Mixed-Humid & Kansas City \\
    \textbf{4} & 5A & Cold-Humid & Omaha \\
    \hline
\end{tabular}
\end{table}

\subsection{State Space and Hyperparameters}

The state space is what the RL agent observes for each control step. The CityLearn environment allows for a total of 27 observations per building that may be passed to the agent. The final state design used in the submission was determined by utilising a combination of expert assessment and trial and error and is outlined in Table \ref{table:3}. A centralised solution is presented here with one agent, which has complete oversight of all nine buildings. The SAC RL algorithm used has several key hyperparameters and the values used in the submission are detailed in Table \ref{table:4}. Note that some parameters were modified in the deployment (evaluation) phase to allow the agent to adapt to new environments (such as a new climate), whilst also retaining the initial weights of the pre-trained agent. 

\begin{table}[!htbp]
\small
\caption{State Space used for Centralised Agent Case}
\label{table:3}
\centering
\begin{tabular}{c p{5cm}} 
 \hline
 State Variable & Description \\ [0.5ex] 
 \hline
 month & month of timestep (1-12) \\ 
 day & day of timestep (1-7) \\
 hour & hour of the day (1-24) \\
 t$_{out}$ & Outside drybulb temperature ($C$) \\
 direct\_solar\_rad & Direct solar radiation ($W/m^2$) \\ [1ex] 
 non\_shiftable\_load$_{n}$ & Non-shiftable electricity load of Building $n$ \\ [1ex] 
 solar\_gen$_{n}$ & Solar generation of Building $n$ (if PV present) \\ [1ex] 
 cooling\_storage\_soc$_{n}$ & State of charge of cooling storage of Building $n$ \\ [1ex] 
 dhw\_storage\_soc$_{n}$ & State of charge of DHW storage of Building $n$ (if present) \\ [1ex] 
 \hline
\end{tabular}
\end{table}

\begin{table}[!htbp]
  \small
  \caption{Hyperparameters for Training and Evaluation}
  \label{table:4}
  \begin{tabular}{cp{2.5cm}cc}
    \toprule
    Symbol & Description & Training Value & Evaluation Value\\
    \midrule
     & Replay buffer size & $2x10^6$ & $2x10^6$ \\
     & Minibatch size & $1024$ & $64$ \\
    $\gamma$ & Discount factor & $0.9$ & $0.9$ \\
    $\alpha$ & Reward temperate parameter & $0.2$ & $0.2$ \\
     & Update interval & $168$ & $168$ \\
     & Learning rate & \num{5e-4} & \num{1e-4} \\
    $\tau$ & Target smoothing coefficient & \num{3e-3} & \num{3e-3} \\
     & Hidden layer size & $256$ & $256$ \\
  \bottomrule
\end{tabular}
\end{table}

\subsection{Reward Function}

The reward function was designed based on a virtual price signal (penalising peak consumption) together with manual reward shaping to incentivise charging during the night and discharging during the day. The reward (R) function used is shown in Eq. 1 where $\beta$ is a weighting coefficient (with a value of 0.005), N is the number of buildings in the district, $e_{total}$ is the total district electricity consumption and $e_{i}$ is the building $i$ electricity consumption. The final reward value was also scaled and clipped to be in the range of -1 to 1.

\begin{equation}
  R = \sum_{j=1}^{N}(\beta*e_{total}*e_{i}) + R_{night} + R_{day}
\end{equation}
where:
\[
    R_{night} = 
\begin{cases}
    1000,& \text{if } 10 pm \leq hour \leq 12 pm \text{ AND } mean(actions) > 0.1\\
    -1000,& \text{if } 10 pm \leq hour \leq 12 pm \text{ AND } mean(actions) < 0\\
    0,              & \text{otherwise}
\end{cases}
\]
\[
    R_{day} = 
\begin{cases}
    -1000,& \text{if } 12 pm \leq hour \leq 08 pm \text{ AND } mean(actions) > 0\\
    0,              & \text{otherwise}
\end{cases}
\]

\section{Results}

\begin{figure}[!htb]
  \centering
  \includegraphics[scale=0.34]{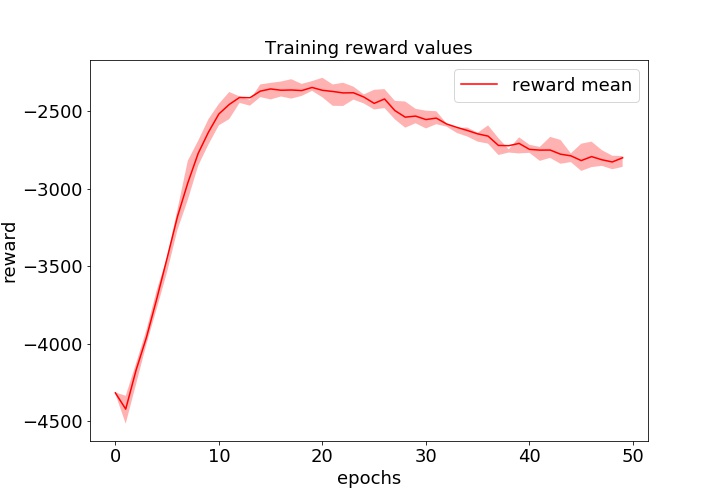}
  \includegraphics[scale=0.34]{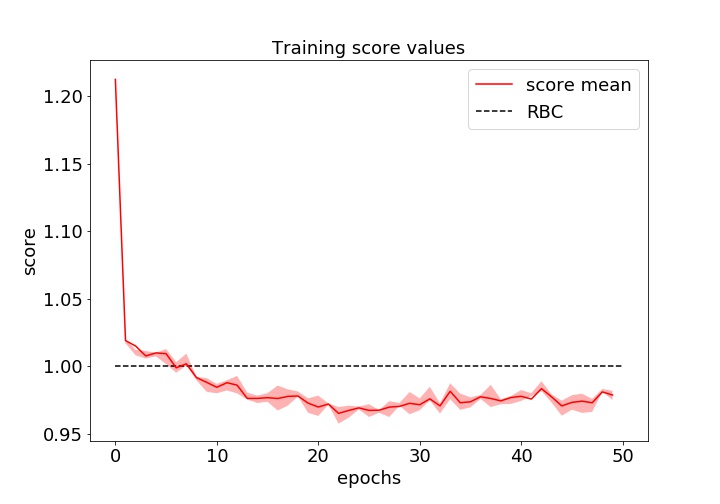}
  \caption{Training results for Climate Zone 1 (reward and cost objectives as function of training episodes).}
  \Description{Graphs showing the score and reward values at each step of the training.}
  \label{fig:rewards_scores}
\end{figure}

\begin{figure*}[!htb]
  \centering
  \includegraphics[scale=0.5]{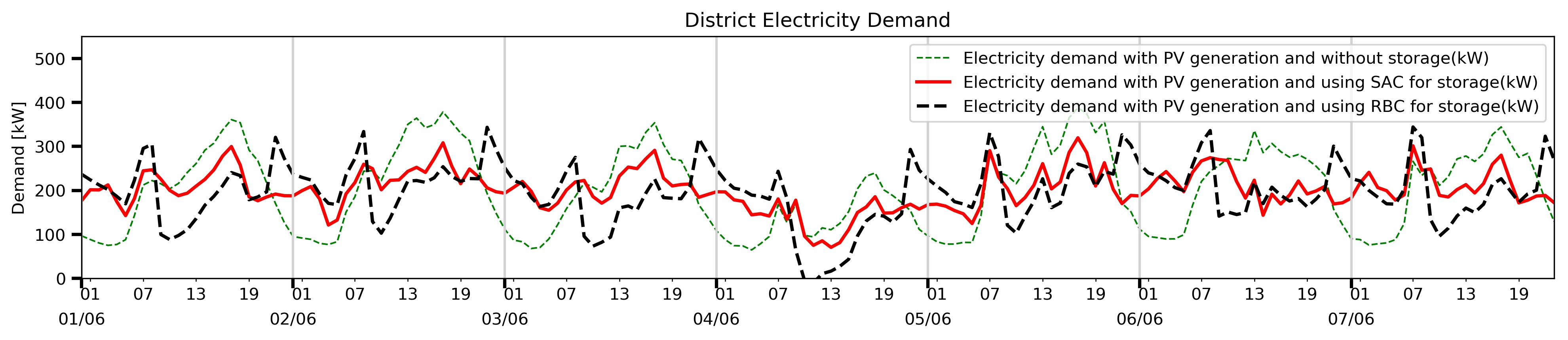}
  \caption{A comparison of district electricity consumption for (i) no load shifting, (ii) predefined RBC and (iii) SAC RL.}
  \label{fig:district_consumption}
\end{figure*}

The centralised SAC RL Agent shows promising performance applied to the district DSM problem. Within 10 episodes of training on the dataset for climate zone 1 (see Fig. \ref{fig:rewards_scores}), the agent realises an improved multi-objective cost function (as defined in Section 3.1) as compared to the RBC baseline (manually predefined controller). Note that the cost function is computed relative to the RBC baseline, i.e., a cost function of less than 1 is considered to be an improved performance over the baseline and a cost function of greater than 1 is considered to be a poorer performance compared to the baseline. When this pre-trained agent is evaluated (i.e., deployed) for the same climate zone, it produces an improvement of 10.7\% (a score of 0.893) over the RBC baseline (see Figure \ref{fig:district_consumption} for the district electricity consumption profile and Table \ref{tab:evaluation} for the scores). This table also shows the ability of the RL agent to generalise and adapt to new climatic conditions, and although they suffered a performance drop, still managed to outperform the manually tuned RBC.

The Challenge dataset scores showed that the agent was further generalisable to unseen data (see Table \ref{tab:evaluation}) featuring both different building properties and climates. Overall, an average improvement of 3.3\% was seen over the manually tuned RBC over the four different climate zones tested. Whilst these improvements are modest, they are promising given the adaptability over a range of buildings and climates and the limited information required for the state space.

\begin{table*}[t]
\small
\caption{Evaluation Results on the Design \& Challenge Dataset}
\label{tab:evaluation}
\textit{Note: The values presented below are relative to the baseline predefined RBC controller which has a score of 1. The lower the score,\\ the better the performance of the RL agent.}
\begin{tabular}{p{1cm}p{2cm}p{2cm}p{2cm}p{2cm}p{2cm}p{2cm}}
\hline
Climate Zone & Ramping & 1-Load Factor & Avg. Daily Peak & Peak Demand & Net Elec. Consumption & Avg. Score \\ \hline
\multicolumn{7}{c}{\textbf{Design Dataset}} \\
\textbf{1} & 0.735 & 0.881 & 0.849 & 0.986 & 1.014 & 0.893 \\
\textbf{2} & 0.777 & 1.017 & 0.940 & 1.187 & 1.018 & 0.988 \\
\textbf{3} & 0.810 & 0.983 & 0.985 & 1.077 & 1.019 & 0.975 \\
\textbf{4} & 0.789 & 0.983 & 0.959 & 1.004 & 1.014 & 0.950 \\ \hline
\multicolumn{1}{l}{} & \multicolumn{1}{l}{} & \multicolumn{1}{l}{} & \multicolumn{1}{l}{} & \multicolumn{1}{l}{} & Avg. Score & 0.952 \\
\hline
\multicolumn{7}{c}{\textbf{Challenge Dataset}} \\
\textbf{1} & 0.779 & 1.014 & 0.982 & 1.131 & 1.015 & 0.984 \\
\textbf{2} & 0.780 & 0.980 & 0.959 & 0.999 & 1.013 & 0.946 \\
\textbf{3} & 0.812 & 0.960 & 0.939 & 1.083 & 1.018 & 0.962 \\
\textbf{4} & 0.860 & 0.996 & 0.991 & 1.013 & 1.017 & 0.976 \\ \hline
\multicolumn{1}{l}{} & \multicolumn{1}{l}{} & \multicolumn{1}{l}{} & \multicolumn{1}{l}{} & \multicolumn{1}{l}{} & Avg. Score & \textbf{0.967} \\
\hline
\end{tabular}
\end{table*}

\normalsize

\section{Conclusions and Further Work}

In this paper, a centralised `Soft Actor Critic' reinforcement learning agent, capable of handling continuous action spaces, is proposed for the district demand side management problem and the performance of the agent applied to the CityLearn challenge is outlined. The agent was able to secure second place in the competition achieving an average score of 0.967 over the challenge dataset featuring different buildings and climates when compared to the reference manually tuned rule-based controller. This highlights the potential of deep reinforcement learning as a plug-and-play style controller for district level demand side management of buildings. Limitations include the manual reward shaping applied which perhaps limits the generalisation ability of the RL agent to districts with significantly different demand profiles. Given the centralised agent with oversight of all buildings, it is not known how the computational requirements and performance would scale over a larger number buildings. A further limitation of CityLearn is that the cooling load is precomputed and hence currently does not support thermal comfort considerations and utilisation of the passive thermal mass for load shifting. Future work aims to further the robustness of the RL agent through reducing the amount of manual reward shaping applied and testing the performance of the algorithm for different hyperparameters. The addition of further energy systems such as batteries and electric vehicles to CityLearn will also be considered.

\begin{acks}
The authors gratefully acknowledge that their contribution emanated from research supported by Science Foundation Ireland under the SFI Strategic Partnership Programme Grant Number SFI/15/SPP/E3125.
\end{acks}

\bibliographystyle{ACM-Reference-Format}
\bibliography{RLEM_anjukan}

\end{document}